\pdfoutput=1

\documentclass[11pt]{article}

\usepackage{ACL2023}

\usepackage{times}
\usepackage{latexsym}

\usepackage[T1]{fontenc}

\usepackage[utf8]{inputenc}

\usepackage{microtype}

\usepackage{inconsolata}

\usepackage{graphicx}
\usepackage{multirow}
\usepackage{makecell}
\usepackage{booktabs}
\usepackage{enumitem}
\usepackage{amssymb}
\usepackage{xcolor}
\usepackage{subcaption}
\definecolor{dark-green}{rgb}{0.31, 0.47, 0.26}
\definecolor{dark-red}{rgb}{0.81, 0.09, 0.13}

\usepackage{soul}
\usepackage{amsmath}

\setuldepth{Berlin}
\usepackage{tabularx}
\newcolumntype{x}[1]{>{\arraybackslash\hspace{0pt}}m{#1}}

\usepackage{pifont}

\newcommand{\paranmt}{\textsc{ParaNMT}}
\newcommand{\parabanka}{\textsc{ParaBank1}}
\newcommand{\parabankb}{\textsc{ParaBank2}}
\newcommand{\paraamr}{\textsc{ParaAMR}}

\title{ParaAMR: A Large-Scale Syntactically Diverse  Paraphrase Dataset by AMR Back-Translation}

\author{
  Kuan-Hao Huang$^{\dagger}$ \ \ Varun Iyer$^{\oplus}$ \ \ I-Hung Hsu$^{\diamond}$ \\ 
  {\bf Anoop Kumar$^{\ddagger}$  \ \ Kai-Wei Chang$^{\dagger}$$^{\ddagger}$ \ \ Aram Galstyan$^{\ddagger}$$^{\diamond}$} \vspace{0.2em}\\
  $^{\dagger}$University of California, Los Angeles \ \ $^{\oplus}$University of Illinois Chicago \\
  $^{\diamond}$Information Science Institute, University of Southern California \ \
  $^{\ddagger}$Amazon Alexa AI\vspace{0.2em}\\
  \texttt{\{khhuang, kwchang\}@cs.ucla.edu}, \ \ \texttt{viyer9@uic.edu} \\ 
  \texttt{ihunghsu@isi.edu}, \ \ \texttt{\{anooamzn, argalsty\}@amazon.com} \\
}

\begin{document}
\maketitle
\begin{abstract}

Paraphrase generation is a long-standing task in natural language processing (NLP).
Supervised paraphrase generation models, which rely on human-annotated paraphrase pairs, are cost-inefficient and hard to scale up.
On the other hand, automatically annotated paraphrase pairs (e.g., by machine back-translation), usually suffer from the lack of syntactic diversity --- the generated paraphrase sentences are very similar to the source sentences in terms of syntax.
In this work, we present \paraamr{}, a large-scale \emph{syntactically diverse} paraphrase dataset created by abstract meaning representation back-translation.
Our quantitative analysis, qualitative examples, and human evaluation demonstrate that the paraphrases of \paraamr{} are syntactically more diverse compared to existing large-scale paraphrase datasets while preserving good semantic similarity.
In addition, we show that \paraamr{} can be used to improve on three NLP tasks: learning sentence embeddings, syntactically controlled paraphrase generation, and data augmentation for few-shot learning.
Our results thus showcase the potential of \paraamr{} for improving various NLP applications.

\end{abstract}

\section{Introduction}
\label{sec:intro}

Paraphrase generation is a long-standing task in natural language processing (NLP) \cite{McKeown83rules,Barzilay03lattice,Kaucha06thres2}.
It has been applied to various downstream applications, such as question answering \cite{Yu18appqa}, chatbot engines \cite{Yan16appcb}, creative generation \cite{Tian21creativegen}, and improving model robustness \cite{Huang21synpg}.
Most existing paraphrase generation models require a large amount of annotated paraphrase pairs \cite{Li19seq2seq5,Gupta18seq2seq6,Kumar20sgcp}.
Since human-labeled instances are expensive and hard to scale up \cite{Dolan04mrpc,Madnani12pan,Iyer17quora}, recent research has explored the possibility of generating paraphrase pairs automatically.
One popular approach is back-translation \cite{Gimpel18paranmt,Hu19parabank1,Hu19parabank2}, which generates paraphrases of a source sentence by translating it to another language and translating back to the original language.
Although back-translation creates large-scale automatically annotated paraphrase pairs, the generated paraphrases usually suffer from the lack of syntactic diversity --- they are very similar to the source sentences, especially in syntactic features.
Consequently, supervised paraphrase models trained with those datasets are also limited in their ability to generate syntactically diverse paraphrases.
Furthermore, not all words can be perfectly translated into another language. As we will show in Section~\ref{sec:quali}, this mismatch may produce subpar paraphrases.

\begin{figure*}[t]
	\centering
    \includegraphics[width=.99\textwidth]{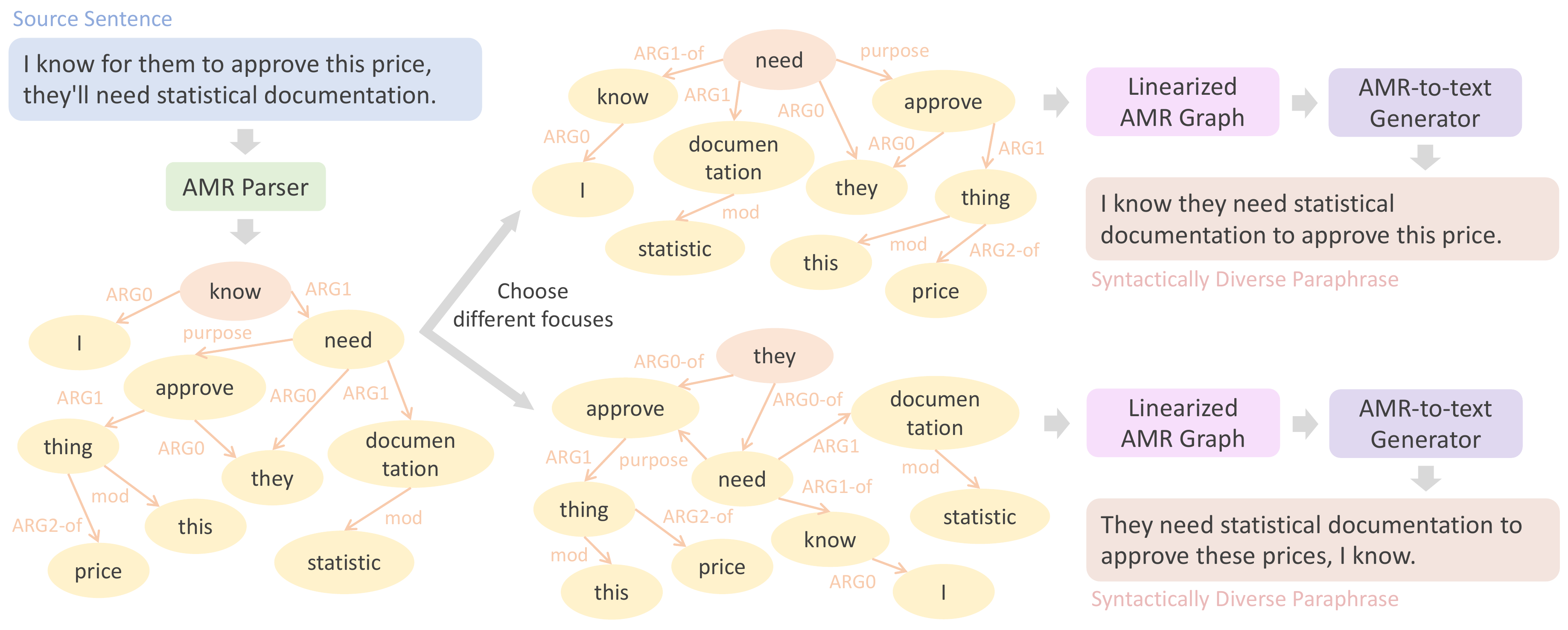}
    \caption{The overall framework to construct \paraamr{} based on AMR back-translation. We encode a source sentence to an AMR graph, modify the \emph{focus} of the AMR graph, linearize the modified AMR graph, and finally decode the linearized graph to a syntactically diverse paraphrase.}
    \label{fig:flow}
\end{figure*}

In this work, we leverage abstract meaning representation (AMR) \cite{Banarescu13amr} to generate syntactically diverse paraphrase pairs.
We present \paraamr{}, a large-scale syntactically diverse paraphrase dataset based on AMR back-translation. 
As illustrated by Figure~\ref{fig:flow}, our approach works by encoding a source sentence to an AMR graph, modifying the \emph{focus} of the AMR graph that represents the main assertion, linearizing the modified AMR graph, and finally decoding the linearized graph back to a sentence.
Since the new sentence shares the same AMR graph structure as the source sentence, it preserves similar semantics to the source sentence. At the same time, the change of \emph{focus} makes the new main assertion different from that source sentence. 
When linearizing the AMR graph, a different concept will be emphasized at the beginning of the string.
Therefore, the decoded sentence may have a much different syntax from the source sentence.

Our quantitative analysis (Section~\ref{sec:quant}) and qualitative examples (Section~\ref{sec:quali}) show that the paraphrases of \paraamr{} are syntactically more diverse than existing datasets \cite{Gimpel18paranmt,Hu19parabank1,Hu19parabank2}, while at the same time preserving good semantic similarity between paraphrased sentences. 
In addition, our human evaluation results (Section~\ref{sec:human}) confirm that \paraamr{} is indeed more syntactically diverse than prior datasets.
To showcase the benefits of syntactically diverse paraphrases, we conduct experiments on three downstream tasks: learning sentence embeddings (Section~\ref{sec:sts}), syntactically controlled paraphrase generation (Section~\ref{sec:scpg}), and data augmentation for few-shot learning (Section~\ref{sec:few}).
We observe that models trained on \paraamr{} achieve better performance on all three downstream tasks compared to other datasets,
thus indicating its potential value for various NLP applications.\footnote{Our proposed dataset is available at \url{https://github.com/uclanlp/ParaAMR}.}

\section{Related Work}
\label{sec:related}

\paragraph{Paraphrase generation and datasets.}
Traditional paraphrase generation models are usually based on hand-crafted rules, including 
rule-based methods \cite{McKeown83rules}, thesaurus-based methods \cite{Bolshakov04thres,Kaucha06thres2}, and lattice matching methods \cite{Barzilay03lattice}.
In recent years, different neural models have been proposed for paraphrase generation \cite{Prakash16seq2seq1,Lapata17seq2seq2,Cao17seq2seq3, Egonmwan19seq2seq4,Li19seq2seq5,Gupta18seq2seq6,Zhang19sivae,Roy19vqvae,Iyyer18scpn,Huang21synpg}. 
Some advanced techniques are proposed as well, such as multi-round generation \cite{Lin21multiround}, reinforcement-learning-based  paraphrasing~\cite{Liu20rl1}, and prompt-tuning \cite{Chowdhury22prompt}.
To properly train those neural models, however,  we needs a large corpus of annotated paraphrase pairs. Most existing paraphrase datasets and related resources, such as MRPC \cite{Dolan04mrpc}, PAN \cite{Madnani12pan}, PPDB \cite{Ganitkevitch13ppdb}, and Quora \cite{Iyer17quora}, have limited scale. Therefore, 
researchers have focused on automatically generating large-scale paraphrase corpora. One notable example is \paranmt{} \cite{Gimpel18paranmt}, which is created by machine back-translation --- translating texts to another language and translating them back to the original language. 

\paragraph{Syntactically diverse paraphrase generation.}
Another line of research focuses on diversifying the generated paraphrases in terms of syntax.
This includes sampling from latent spaces \cite{Roy19vqvae,Zhang19sivae,Cao20latent2}, controlling word order \cite{Goyal20order1}, and controlling syntax \cite{Iyyer18scpn,Cao19ct1,Kumar20sgcp,Huang21synpg,Sun21aesop,Huang22amrpg,Lee22ct2}.
Although they can diversify the generated paraphrases based on different model designs, those models are still limited due to the lack of diversity in existing large-scale paraphrase datasets.
Some works propose large-scale diverse paraphrases by considering different decoding methods during back-translation, including lexical constraints \cite{Hu19parabank1} and cluster-based constrained sampling \cite{Hu19parabank2}.
Although increasing the lexical diversity, the syntactic diversity of their datasets is still limited.

\paragraph{Text-to-AMR parsing.}
Abstract meaning representation (AMR) \cite{Banarescu13amr} is designed for capturing abstract semantics.
Since it offers benefits to many NLP tasks, several works focus on parsing  AMR from texts. 
Transition-based methods maintain a stack and a buffer for parsing AMR \cite{Wang15amrtrans1,Damonte17amrtrans2,Ballesteros17amrtrans3,Vilares18amrtrans4,Naseem19amrtrans5}.
Graph-based approaches extract AMR based on graph information \cite{Zhang19amrg1,Zhang19amrg2,Cai20amrg3,Cai20amrg4}.
Sequence-to-sequence approaches directly linearize AMR and train end-to-end models to produce AMR \cite{Konstas17amrp1,Noord17amrp2,Gildea17amrp3,Ge19amrp4}.

\paragraph{AMR-to-Text generation.}
Generating texts from AMR graphs is a popular research direction as well.
Most existing approaches can be grouped into two categories.
The first group is based on structure-to-text methods, where they build graphs to capture the structural information \cite{amrgg1,amrgg2,amrgg3,amrgg4,amrgg5,Wang20amrgg6}.
The second group is based on sequence-to-sequence methods \cite{Konstas17amrgs1,Ribeiro21amrdecoder}, where they treat AMR as a string and train end-to-end models.

\section{\paraamr{}}
\label{sec:data}
We propose \paraamr{}, a large-scale syntactically diverse paraphrase dataset.
Figure~\ref{fig:flow} illustrates the overall framework to construct \paraamr{} by AMR back-translation.
In summary, we encode a source sentence to an AMR graph, modify the \emph{focus} of the AMR graph (see Section~\ref{sec:amrtotext}), linearize the modified AMR graph, and finally decode the linearized graph to a syntactically diverse paraphrase.
We describe the details in the following.

\subsection{Data Source}
In order to fairly compare with prior works \cite{Gimpel18paranmt,Hu19parabank1,Hu19parabank2}, we choose the same Czech–English dataset \cite{Bojar16czen} as our data source.
Specifically, we directly use the English source sentences from the previous dataset \cite{Hu19parabank2} as the source sentences for AMR back-translation.
It is worth noting that our proposed method is not limited to this dataset but can be applied to any general texts for constructing syntactically diverse paraphrases.

\subsection{Translating Texts to AMR Graphs}
We use a pre-trained AMR parser to encode source sentences to AMR graphs.
Specifically, we consider SPRING \cite{Bevilacqua21amrparse1}, a BART-based \cite{Lewis20bart} AMR parser trained on AMR 3.0 annotations\footnote{\url{https://catalog.ldc.upenn.edu/LDC2020T02}} and implemented by amrlib.\footnote{\url{https://github.com/bjascob/amrlib}}
As illustrated by Figure~\ref{fig:flow}, each source sentence will be encoded to an AMR graph, which is a directed graph that has each node represents a semantic concept (e.g., \emph{know}, \emph{need}, and \emph{they}) and each edge describe the semantic relations between two concepts (e.g., \emph{ARG0}, \emph{ARG1-of}, and \emph{mod}) \cite{Banarescu13amr}.

An AMR graph aims at capturing the meaning of a sentence while abstracting away syntactic, lexical, and other features. Each AMR graph has a \emph{focus}, which is the root node of the graph, to represent the main assertion.
For example, the focus of the AMR graph extracted from the source sentence in Figure~\ref{fig:flow} is \emph{know}.
Most of the time, the focus will be the main verb; however, it actually can be any concept node.

\subsection{Translating AMR Graphs to Texts}
\label{sec:amrtotext}
Usually, syntactically different sentences with similar meanings have similar \emph{undirected} AMR graph structures and differ only in their focuses and the directions of edges.
We plan to use this property to construct syntactically diverse paraphrases of a source sentence.

\paragraph{Changing the focus of an AMR graph.}
After extracting the AMR graph from a source sentence, we construct several new graphs by changing the \emph{focus}.
More precisely, we randomly choose a node as the new focus and reverse all the incoming edges for that node.
For instance, in Figure~\ref{fig:flow}, when we choose \emph{need} as the new focus, the incoming edge from \emph{know} is reversed, and its edge label changes from \emph{ARG1} to \emph{ARG1-of}.
Similarly, when we choose \emph{they} as the new focus, the incoming edge from \emph{need} and \emph{approve} are reversed, and their edge labels change from \emph{ARG0} to \emph{ARG0-of}.
Sometimes, to maintain a tree-like graph, some outgoing edges of the original focus node will be reversed as well (e.g., the edge between \emph{know} and \emph{need} is reversed when we choose \emph{they} as the new focus).
It is worth noting that when the focus changes, the \emph{undirected} AMR graph structure remains the same, meaning that the new AMR graph preserves a similar abstract meaning to the old one.
We implement the process of AMR re-focusing by the PENMAN package \cite{Goodman20penman}.\footnote{\url{https://github.com/goodmami/penman}}

\paragraph{Linearizing AMR graph.}
After constructing several new graphs from the original AMR graph, we linearize the new graphs with the new focus (root node).
This is done by traversing the AMR graph starting from the new focus node with a depth-first-search algorithm and converting it to the PENMAN notation.
For example, the AMR graph with the focus being \emph{need} can be linearized in the following format:
\\[8pt]
\fbox{\begin{minipage}{0.47\textwidth}
\texttt{(z3 / need \\
\hspace*{2em}:ARG1-of (z1 / know \\
\hspace*{4em}:ARG0 (z2 / i)) \\
\hspace*{2em}:ARG0 (z4 / they) \\ 
\hspace*{2em}:ARG1 (z5 / documentation \\
\hspace*{4em}:mod (z6 / statistic)) \\
\hspace*{2em}:purpose (z7 / approve \\
\hspace*{4em}:ARG0 z4 \\
\hspace*{4em}:ARG1 (z8 / thing \\
\hspace*{6em}:ARG2-of (z9 / price) \\
\hspace*{6em}:mod (z10 / this))))}
\end{minipage}}
\\[8pt]
Similarly, the AMR graph with the focus node \emph{they} can be linearized in the following format:
\\[8pt]
\fbox{\begin{minipage}{0.47\textwidth}
\texttt{(z4 / they \\
\hspace*{2em}:ARG0-of (z3 / need \\
\hspace*{4em}:ARG1 (z5 / documentation \\
\hspace*{6em}:mod (z6 / statistic)) \\
\hspace*{4em}:purpose (z7 / approve \\
\hspace*{6em}:ARG0 z4 \\
\hspace*{6em}:ARG1 (z8 / thing \\
\hspace*{8em}:ARG2-of (z9 / price) \\
\hspace*{8em}:mod (z10 / this))) \\
\hspace*{4em}:ARG1-of (z1 / know \\
\hspace*{6em}:ARG0 (z2 / i))))}
\end{minipage}}
\\[8pt]

\paragraph{Decoding AMR graph to texts.}
 We use a T5-based pre-trained AMR-to-text generator \cite{Ribeiro21amrdecoder} to translate the linearized graphs back to sentences.
Since the generated sentences share the same \emph{undirected} AMR graph as the source sentence, they should have similar meanings and thus can be considered as paraphrases of the source sentence. In addition, we observe that the pre-trained AMR-to-text generator tends to emphasize the focus node of an AME graph at the beginning of the generated sentence. Therefore, the generated sentences from the linearized graphs with different focuses are very likely syntactically different from the source sentence.

\subsection{Post-Processing}
We notice that not all nodes are appropriate to be the focus.
Choosing inappropriate nodes as the focus might generate paraphrases that are not grammatically fluent or natural.
To avoid this situation, we use perplexity to filter out bad paraphrases. Specifically, we consider the GPT-2 model \cite{Radford2019gpt2} implemented by HuggingFace's Transformers \cite{Wolf20trasnformers} to compute the perplexity of a candidate paraphrase. We found that setting the filtering threshold to 120 is generally good enough, although some downstream applications may need different thresholds.

\begin{table*}[t!]
\small
\centering
\setlength{\tabcolsep}{7pt}
\resizebox{.82\textwidth}{!}{
\begin{tabular}{l|c|c|c}
    \toprule
    Dataset & \#Instances & Avg. \#Para. & Avg. Len. \\
    \midrule
    \paranmt{} \cite{Gimpel18paranmt} & 51,409,584 & 1.00 & 11.90 \\
    \parabanka{} \cite{Hu19parabank1} & 57,065,358 & 4.31 & 12.16 \\
    \parabankb{} \cite{Hu19parabank2} & 19,723,003 & 4.75 & 15.51 \\
    \paraamr{} (Ours) & 15,543,606 & 6.91 & 15.20 \\
    \bottomrule
\end{tabular}}
\caption{Basic statistics of \paranmt{}, \parabanka{}, \parabankb{}, and \paraamr{}.}
\label{tab:stats}
\end{table*}

\begin{table*}[t!]
\centering
\setlength{\tabcolsep}{5pt}
\resizebox{.99\textwidth}{!}{
\begin{tabular}{l|c|cc|cc}
    \toprule
    \multirow{2}{*}{Dataset} & Semantic  & \multicolumn{2}{c|}{Lexical Diversity} & \multicolumn{2}{c}{Syntactic Diversity} \\
    & Similarity ($\uparrow$) & 1 - BLEU ($\uparrow$) & 1 - $\cap/\cup$ ($\uparrow$) & TED-3 ($\uparrow$) & TED-F ($\uparrow$) \\
    \midrule
    \paranmt{} \cite{Gimpel18paranmt} & \textbf{84.28} & 70.71 & 45.78 & 3.28 & 13.94 \\
    \parabanka{} \cite{Hu19parabank1} & 81.77 & 78.19 & 52.59 & 3.59 & 14.53 \\
    \parabankb{} \cite{Hu19parabank2} & 82.50 & \textbf{88.82} & \textbf{59.61} & 4.04 & 17.41 \\
    \paraamr{} (Ours)   & 82.05 & 87.86 & 53.10 & \textbf{5.86} & \textbf{22.07} \\
    \bottomrule
\end{tabular}}
\caption{Paraphrase diversity of different datasets. \paraamr{} is syntactically more diverse than other datasets, while also showing comparable semantic similarity.}
\label{tab:metric}
\end{table*}

\section{Comparison to Prior Datasets}
\label{sec:cmp}

We compare \paraamr{} with the following three datasets.
(1) \paranmt{} \cite{Gimpel18paranmt} create paraphrase pairs by English-Czech-English back-translation.
(2) \parabanka{} \cite{Hu19parabank1} adds lexical constraints during the decoding of back-translation to increase the lexical diversity of generated paraphrases.
(3) \parabankb{} \cite{Hu19parabank2} proposes cluster-based constrained sampling to improve the syntactic diversity of generated paraphrases.

\subsection{Basic Statistics}
Table~\ref{tab:stats} lists the statistics of the \paranmt{}, \parabanka{}, \parabankb{}, and \paraamr{}.
\paraamr{} contains syntactically diverse paraphrases to around 15 million source sentences.
Notice that we consider the same source sentences as \parabankb{}; however, some of the sentences fail to be parsed into ARM graphs.
Therefore, the size of \paraamr{} is slightly smaller than \parabankb{}.
The average length of paraphrases in \paraamr{} is 15.20, which is similar to \parabankb{}.
Each source sentence in \paraamr{} has 6.91 paraphrases on average, which is more than the other three datasets.

\subsection{Quantitative Analysis}
\label{sec:quant}
Following previous work \cite{Hu19parabank2}, we consider the same metrics to analyze semantic similarity, lexical diversity, and syntactic diversity of different paraphrase datasets.
To fairly compare different datasets, we consider only those examples whose source sentences appear in all datasets. 
There are 193,869 such examples in total.
All the following metrics are calculated based on those 193,869 examples.

We use the following metrics to evaluate the semantic similarity of paraphrases:
\begin{itemize}[topsep=3pt, itemsep=0pt, leftmargin=12pt]
    \item \textbf{Semantic similarity measure by SimCS}E: Given two paraphrase sentences, we use the supervised SimCSE model \cite{Gao21simcse} to get the sentence embeddings, and compute the cosine similarity between the two sentence embeddings as the semantic similarity.
\end{itemize}
Following the previous work \cite{Hu19parabank2}, we consider the following automatic metrics for lexical diversity:
\begin{itemize}[topsep=3pt, itemsep=0pt, leftmargin=12pt]
    \item \textbf{1 - BLEU} ($\uparrow$): We compute one minus BLEU score as the diversity score.
    \item \textbf{1 - $\cap/\cup$} ($\uparrow$): We first compute the ratio of the number of shared tokens between the two sentences and the union of all tokens in the two sentences, then use one minus the ratio as the diversity score.
\end{itemize}
We consider the following automatic metrics for syntactic diversity:
\begin{itemize}[topsep=3pt, itemsep=0pt, leftmargin=12pt]
    \item \textbf{TED-3 }($\uparrow$): We first get the constituency parse trees of the two sentences by using the Stanford CoreNLP parser \cite{Manning14corenlp}. Then, we only consider the top-3 layers of trees and compute the tree editing distance as the score.
    \item \textbf{TED-F} ($\uparrow$): We first get the constituency parse trees of the two sentences by using the Stanford CoreNLP parser \cite{Manning14corenlp}. Then, we consider the whole tree and compute the tree editing distance as the score.
\end{itemize}
From Table~\ref{tab:metric}, we conclude that the paraphrases generated by \paraamr{} increase much more syntactic diversity while preserving comparable semantics compared to prior datasets.

\begin{table*}[t!]
\scriptsize
\centering
\aboverulesep = 0.3mm
\belowrulesep = 0.6mm
\setlength{\tabcolsep}{7pt}
\resizebox{.95\textwidth}{!}{
\begin{tabular}{ll}
    \toprule
    Source Sentence & I know for them to approve this price, they'll need statistical documentation. \\
    \midrule
    \paranmt & I know that in order to accept this award, they'll need a statistical analysis. \\
    \midrule
    \multirow{3}{*}{\parabanka} & I know that to accept this prize, they're going to need statistical analysis. \\
    & I know that in order to accept this prize, they're going to need a statistic analysis. \\
    & I know that if they accept this prize, they're gonna need a statistical analysis. \\
    \midrule
    \multirow{3}{*}{\parabankb} & I know that to accept that prize, they're going to need a statistical analysis. \\
    & I know that in order to accept this prize, they will require a statistical analysis. \\
    & I know they'll require statistical analysis to accept that prize. \\
    \midrule
    \multirow{3}{*}{\paraamr} & I know they need statistical documentation to approve this price. \\
    & There is statistic documentation I know they need to approve these prices. \\
    & They need statistical documentation to approve these prices, I know. \\
    \midrule
    \vspace{-0.7em} \\
    \midrule
    Source Sentence & If I wanted to paddle down the river, where's the best place to launch out of? \\
    \midrule
    \paranmt & If I wanted to row down a river, where's the best place to swim? \\
    \midrule
    \multirow{3}{*}{\parabanka} & If I wanted to row down the river, where's the best place to go? \\
    & If I wanted to row down the riverside, where's the best place to go? \\
    & If I wanted to row down the river, where's the best spot to float? \\
    \midrule
    \multirow{1}{*}{\parabankb} & If I want to paddle down the river, what'd be the most perfect spot to set sail? \\
    \midrule
    \multirow{3}{*}{\paraamr} & Where would be best for me to launch if I wanted to paddle down the river? \\
    & It's a river I want to paddle down to, where's the best place to launch? \\
    & Where's my best place to launch if I want to paddle down the river? \\
    \bottomrule
    
\end{tabular}}
\caption{Paraphrases generated by different datasets. The generated paraphrases by \paranmt{}, \parabanka{}, and \parabankb{} usually have similar syntactic structures to the source sentences. In contrast, \paraamr{} generates more syntactically diverse paraphrases.}
\label{tab:para}
\end{table*}

\begin{table*}[t!]
\centering
\setlength{\tabcolsep}{5pt}
\resizebox{.92\textwidth}{!}{
\begin{tabular}{l|cccc|cccc}
    \toprule
    \multirow{2}{*}{Datasets} & \multicolumn{4}{c|}{Semantic Similarity} & \multicolumn{4}{c}{Syntactic Diversity} \\
    & 3(\%) & 2(\%) & 1(\%) & Average & 3(\%) & 2(\%) & 1(\%) & Average \\
    \midrule
    \paranmt{} \cite{Gimpel18paranmt}  
    & 28.7 & 46.7 & 24.6 & \textbf{2.04} & 16.7 & 45.0 & 38.3 & 1.78 \\
    \parabanka{} \cite{Hu19parabank1} 
    & 26.8 & 49.0 & 24.2 & 2.03 & 15.1 & 47.8 & 37.1 & 1.78 \\
    \parabankb{} \cite{Hu19parabank2} 
    & 26.8 & 50.3 & 22.9 & \textbf{2.04} & 14.2 & 51.8 & 34.0 & 1.80 \\
    \paraamr{} (Ours)
    & 26.5 & 47.2 & 26.3 & 2.00 & 18.2 & 53.8 & 28.0 & \textbf{1.90} \\
    \bottomrule
\end{tabular}}
\caption{Human evaluation results. We evaluate semantic similarity and syntactic diversity in a score of three and report the distribution and the average score.}
\label{tab:human}
\vspace{-1em}
\end{table*}

\subsection{Qualitative Examples}
\label{sec:quali}

Table~\ref{tab:para} shows some paraphrases generated by different datasets.
We can observe that prior datasets based on machine back-translation tend to only replace synonyms as paraphrases.
In contrast, \paraamr{} is able to generate paraphrases that have much different word order and syntactic structures compared to the source sentence.
This again showcases the syntactic diversity of \paraamr{}.

In addition, we notice that other datasets may change the meaning of the source sentence (e.g., from \emph{price} to \emph{prize} and from \emph{paddle} to \emph{row}) due to the translation errors between different languages.
\paraamr{}, on the other hand, does not depend on other languages and thus is more reliable.

\subsection{Human Evaluation}
\label{sec:human}
We additionally conduct human evaluations to measure the semantic similarity and the syntactic diversity of different datasets.
We used the Amazon Mechanical Turk\footnote{\url{https://www.mturk.com/}} to conduct the human evaluation.
We randomly sample 300 paraphrases from each dataset, and design questions to measure the semantic similarity and syntactic diversity.

For semantic similarity, we design a 3-point scale question and ask the annotators to answer the question:
\begin{itemize}[topsep=3pt, itemsep=0pt, leftmargin=12pt]
    \item \textbf{Score 3}: The two sentences are paraphrases of each other. Their meanings are near-equivalent.
    \item \textbf{Score 2}: The two sentences have similar meanings but some unimportant details differ.
    \item \textbf{Score 1}: Some important information differs or is missing, which alters the intent or meaning.
\end{itemize}
For syntactic diversity, we design a 3-point scale question and ask the annotators to answer the question:
\begin{itemize}[topsep=3pt, itemsep=0pt, leftmargin=12pt]
    \item \textbf{Score 3}: The two sentences are written in very different ways or have much different sentence structures. (For example, ``\emph{We will go fishing if tomorrow is sunny.}'' and ``\emph{If tomorrow is sunny, we will go fishing}'')
    \item \textbf{Score 2}: Only some words in the two sentences differ. (For example, ``\emph{We will go fishing if tomorrow is sunny.}'' and ``\emph{We are going to go fishing if tomorrow is sunny.}'')
    \item \textbf{Score 1}: The two sentences are almost the same.
\end{itemize}

Appendix~\ref{app:screen} lists more details of human evaluation.
The average scores of human evaluation are shown in Table~\ref{tab:human}.
We observe that \paraamr{} gets a much higher score for syntactic diversity although it has a slightly lower score for semantic similarity.

\section{Applications}
\label{sec:exp}

We focus on three downstream applications of \paraamr{} corpus: learning sentence embeddings (Section~\ref{sec:sts}), syntactically controlled paraphrase generation (Section~\ref{sec:scpg}), and data augmentation for few-shot learning (Section~\ref{sec:few}).
We demonstrate the strength of \paraamr{} and compare with prior datasets: \paranmt{} \cite{Gimpel18paranmt}, \parabanka{} \cite{Hu19parabank1}, and \parabankb{} \cite{Hu19parabank2}.

\subsection{Learning Sentence Embeddings}
\label{sec:sts}

We conduct experiments to show that \paraamr{} is beneficial to learn sentence embeddings because of its syntactic diversity.

\paragraph{Settings.}
We consider the supervised SimCSE \cite{Gao21simcse}, a contrastive learning framework to learn sentence embeddings from (\emph{reference sentence}, \emph{positive sentence}, \emph{negative sentence}) triplets.
We train different SimCSE models with the paraphrase pairs in all four datasets.
Specifically, for each (\emph{source sentence}, \emph{paraphrase sentence}) pair in the dataset, we consider the source sentence as the reference sentence, consider the paraphrase sentence as the positive sentence, and randomly sample one sentence from the dataset as the negative sentence.

\paragraph{Training details.}
We use the script provided by the SimCSE paper\footnote{\url{https://github.com/princeton-nlp/SimCSE}} \cite{Gao21simcse} to train a SimCSE model with the weights initialized by \texttt{bert-base-uncased} \cite{Devlin19bert}.
The batch size is set to 128 and the number of epochs is 3.
We set the learning rate to $10^{-5}$ and set other parameters as the default values from the script.
It takes around 3 hours to train the SimCSE models for a single NVIDIA RTX A6000 GPU with 48GB memory.
We set the perplexity threshold to 110 to filter \paraamr{}.
For each dataset, we train 5 different models with 5 different random seeds and report the average scores.

\paragraph{Evaluation.}
To evaluate the quality of sentence embeddings, we consider sentence textual similarity (STS) tasks from SentEval 2012 to 2016 \cite{Agirre12sts12,Agirre13sts13,Agirre14sts14,Agirre15sts15,Agirre16sts16}.
We consider the script from SentEval\footnote{\url{https://github.com/facebookresearch/SentEval}} and use the learned sentence embeddings to calculate the cosine similarity between two sentences.
We report the average Pearson correlation coefficient and the average Spearman correlation coefficient over all tasks.

\paragraph{Experimental results.}
Table~\ref{tab:sts} lists the average score for STS 2012 to 2016.
We observe that the sentence embeddings learned with \paraamr{} get better scores than other datasets, especially for the Pearson correlation coefficient. 
We hypothesize that the syntactic diversity of \paraamr{} makes the sentence embeddings capture semantics better and reduce the influence of syntactic similarity.

\begin{table}[t!]
\centering
\setlength{\tabcolsep}{9pt}
\resizebox{.48\textwidth}{!}{
\begin{tabular}{l|cc}
    \toprule
    Dataset      & Pearson's r & Spearman's r \\
    \midrule
    \paranmt{}   & 74.38 {\scriptsize $\pm$ 0.70} 
                 & 73.80 {\scriptsize $\pm$ 0.42} \\
    \parabanka{} & 74.80 {\scriptsize $\pm$ 1.33} 
                 & 74.56 {\scriptsize $\pm$ 1.02} \\
    \parabankb{} & 75.39 {\scriptsize $\pm$ 0.29} 
                 & 75.17 {\scriptsize $\pm$ 0.25} \\
    \paraamr{} (ours) & \textbf{77.70} {\scriptsize $\pm$ 0.40} 
               & \textbf{75.72} {\scriptsize $\pm$ 0.43} \\
    \bottomrule
\end{tabular}}
\caption{Results of learning sentence embeddings. We report 5-run average scores for STS 2012 to 2016. \paraamr{} achieves the best performance.}
\label{tab:sts}
\end{table}

\subsection{Syntactically Controlled Paraphrase Generation}
\label{sec:scpg}
We demonstrate that \paraamr{} is better for training a syntactically controlled paraphrase generator.

\paragraph{Settings.}
We consider the same setting as the previous works \cite{Iyyer18scpn,Huang21synpg}, which uses constituency parses as the control signal to train paraphrase generators.
More precisely, the goal is to train a syntactically controlled paraphrase generator with the input being (\emph{source sentence}, \emph{target constituency parse}) pair and the output being a paraphrase sentence with syntax following the target constituency parse.

We consider the SCPN model \cite{Iyyer18scpn}, which is a simple sequence-to-sequence model, as our base model.
We train different SCPN models with different datasets.
for each (\emph{source sentence}, \emph{paraphrase sentence}) pair in the dataset, we treat the paraphrase sentence as the target sentence and use the Stanford CoreNLP toolkit \cite{Manning14corenlp} to extract constituency parse from the paraphrase sentence as the target parse.

\paragraph{Training details.}
Unlike the original SCPN paper \cite{Iyyer18scpn}, which uses LSTM as the base model, we fine-tune the pre-trained \texttt{bart-base} \cite{Lewis20bart} to learn the syntactically controlled paraphrase generator.
The batch size is set to 32 and the number of epochs is 40.
The max lengths for source sentences, target sentences, and target syntax are set to 60, 60, and 200, respectively.
We set the learning rate to $3\times 10^{-5}$ and consider the Adam optimizer without weight decay.
For the beam search decoding, the number of beams is set to 4.
It takes around 12 hours to train the SCPN model for a single NVIDIA RTX A6000 GPU with 48GB memory.
We set the perplexity threshold to 85 to filter \paraamr{}.
For each dataset, we train 5 different models with 5 different random seeds and report the average scores.

\begin{table}[t!]
\centering
\setlength{\tabcolsep}{4pt}
\resizebox{.48\textwidth}{!}{
\begin{tabular}{l|ccc}
    \toprule
    Dataset      & Quora & MRPC & PAN \\
    \midrule
    \paranmt{}   & 47.38 {\scriptsize $\pm$ 0.39} 
                 & 45.24 {\scriptsize $\pm$ 0.61} 
                 & 39.45 {\scriptsize $\pm$ 0.50} \\
    \parabanka{} & 46.21 {\scriptsize $\pm$ 0.26} 
                 & 44.52 {\scriptsize $\pm$ 0.18} 
                 & 39.85 {\scriptsize $\pm$ 0.11} \\
    \parabankb{} & 46.86 {\scriptsize $\pm$ 0.45} 
                 & 45.17 {\scriptsize $\pm$ 0.39} 
                 & 40.20 {\scriptsize $\pm$ 0.56} \\
    \paraamr{} (ours) & \textbf{48.50} {\scriptsize $\pm$ 0.11} 
               & \textbf{47.38} {\scriptsize $\pm$ 0.19} 
               & \textbf{40.30} {\scriptsize $\pm$ 0.10} \\
    \bottomrule
\end{tabular}}
\caption{Results of syntactically controlled paraphrase generation. We report 5-run average BLEU scores for Quora, MRPC, and PAN. \paraamr{} performs the best.}
\label{tab:scpg}
\end{table}

\paragraph{Evaluation.}
We consider three human-annotated paraphrase datasets: Quora \cite{Iyer17quora}, MRPC \cite{Dolan04mrpc}, and PAN \cite{Madnani12pan}, as the testing datasets.
Specifically, we use the testing examples provided by previous work\footnote{\url{https://github.com/uclanlp/synpg}} \cite{Huang21synpg} and calculate the BLEU score between the ground-truth and the generated output as the evaluation metric.

\paragraph{Experimental results.}
Table~\ref{tab:scpg} shows the results of syntactically controlled paraphrase generation.
The paraphrase generator trained with \paraamr{} performs significantly better than others.
We believe this is because \paraamr{} provides several syntactically different paraphrases for one source sentence, therefore helping the paraphrase generator to better learn the association between parse and words.

\subsection{Data Augmentation for Few-Shot Learning}
\label{sec:few}
Finally, we show that \paraamr{} is helpful to generate augmented data for few-shot learning.

\paragraph{Settings.}
We choose the following three classification tasks from GLUE \cite{Wang19glue}: MRPC, QQP, and RTE.
We randomly sample 15 and 30 instances to train classifiers as the few-shot baseline.
Since most tasks in GLUE do not provide the official test labels, we randomly sample 1/3 of instances from the dev set as the internal dev set and use the rest 2/3 instances as the testing set.

For each dataset, we use the learned syntactically controlled paraphrase generators from Section~\ref{sec:scpg} to generate three augmented examples with different parses for each training instance.
More specifically, we first use the pre-trained SCPN model \cite{Iyyer18scpn} to generate the full parse trees from the following three parse templates: \texttt{(ROOT(S(NP)(VP)(.)))}, \texttt{(ROOT(S(VP)(.)))}, and \texttt{(ROOT(NP(NP)(.)))}.
Then we use the generated full parse trees as the target parse for the syntactically controlled paraphrase generator.
Finally, we train a classifier with the original 30 training instances and the augmented examples.

\begin{table}[t!]
\centering
\setlength{\tabcolsep}{9pt}
\resizebox{.45\textwidth}{!}{
\begin{tabular}{l|ccc}
    \toprule
    Dataset      & MRPC & QQP & RTE \\
    \midrule
    \multicolumn{4}{c}{15-Shot Learning} \\
    \midrule
    15-Shot Baseline & 59.93 & 63.18 & 54.05 \\
    \paranmt{}   & 49.26 & 63.54 & \textbf{55.68} \\
    \parabanka{} & 59.56 & 63.72 & 54.59 \\
    \parabankb{} & 58.46 & 63.54 & 54.05 \\
    \paraamr{} (ours) & \textbf{62.87} & \textbf{64.08} & 52.97 \\
    \midrule
    \multicolumn{4}{c}{30-Shot Learning} \\
    \midrule
    30-Shot Baseline & 68.38 & 64.93 & 54.51 \\
    \paranmt{}   & 67.65 & 66.20 & 52.71 \\
    \parabanka{} & 64.46 & 64.86 & 53.79 \\
    \parabankb{} & 68.38 & 64.91 & 54.15 \\
    \paraamr{} (ours) & \textbf{69.36} & \textbf{67.03} & \textbf{55.60} \\
    \bottomrule
\end{tabular}}
\caption{\paraamr{} has better performance of few-shot learning with data augmentation.}
\label{tab:few}
\end{table}

\paragraph{Training details.}
For the few-shot classifiers, we fine-tune \texttt{bert-base-uncased} \cite{Devlin19bert}.
We set the batch size to 8, set the learning rate to $10^{-4}$, and set the number of epochs to 20.
We consider Adam optimizer with weight decay being $10^{-5}$.
It takes around 5 minutes to train a few-shot classifier for a single NVIDIA RTX A6000 GPU with 48GB memory.
We set the perplexity threshold to 110 to filter \paraamr{}.

\paragraph{Experimental results.}
The results in Table~\ref{tab:few} demonstrate that leveraging \paraamr{} for data augmentation in few-shot learning scenarios leads to consistently better results compared to other paraphrasing corpora. This observation, combined with the two previous experiments, showcases the potential value of \paraamr{} for various NLP applications.

\section{Conclusion}

In this work, we present \paraamr{}, a large-scale syntactically diverse paraphrase dataset created by AMR back-translation.
Our quantitative analysis, qualitative examples, and human evaluation demonstrate that the paraphrases of \paraamr{} are more syntactically diverse than prior datasets while preserving semantic similarity.
In addition, we conduct experiments on three downstream tasks, including learning sentence embeddings, syntactically controlled paraphrase generation, and data augmentation for few-shot learning, to demonstrate the advantage of syntactically diverse paraphrases.

\section*{Acknowledgments}
We thank anonymous reviewers for their helpful feedback. We thank Amazon Alexa AI and the UCLA-NLP group for the valuable discussions and comments. 
\section*{Limitations}

Our goal is to demonstrate the potential of using AMR to generate syntactically diverse paraphrases.
Although we have shown the strength of diverse paraphrases, there are still some limitations.
First, our proposed techniques are strongly based on the quality of pre-trained text-to-AMR parsers and pre-trained AMR-to-text generators.
If we cannot get a strong pre-trained text-to-AMR parser and a pre-trained AMR-to-text generator, the generated paraphrases might not have good quality.
Second, one step in our proposed framework is modifying the root node of the AMR graph and therefore changing the focus of the AMR graph. 
However, not all nodes can be good root nodes to generate appropriate paraphrases. 
Some of them can be not fluent and much different from natural sentences.
Although we use perplexity to filter out those paraphrases, there must be some imperfect paraphrases remaining.
This partially affects the semantic scores of \paraamr{}.
Nevertheless, we still show that the current quality of \paraamr{} is good enough to improve at least three NLP tasks.

\section*{Broader Impacts}
Our dataset construction process relies on a pre-trained AMR-to-text generator. 
It is known that the models trained with a large text corpus may capture the bias reflecting the training data. 
Therefore, it is possible that \paraamr{} contains offensive or biased content learned from the data. 
We suggest to carefully examining the potential bias before applying our dataset to any real-world applications.

\bibliography{custom}
\bibliographystyle{acl_natbib}

\clearpage
\appendix
\appendix

\section{Details of Human Evaluation}
\label{app:screen}

\begin{figure}[h!]
	\centering
    \includegraphics[width=.99\textwidth]{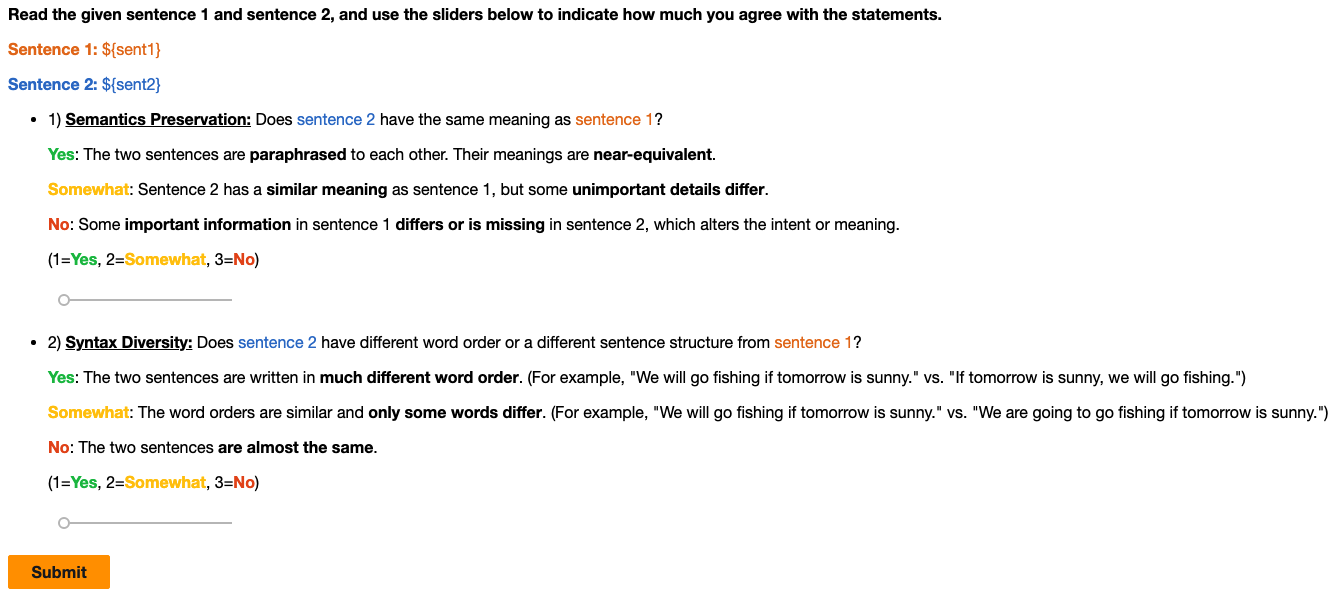}
    \caption{Screenshot of human evaluation instructions.}
    \label{fig:screen}
\end{figure}

We use the template shown in Figure~\ref{fig:screen} to conduct the human evaluation.  
We sampled 300 paraphrases from \paranmt{}, \parabanka{}, \parabankb{}, and \paraamr{} that share the same source sentences for human evaluation.

For each paraphrase pair, we ask three MTurkers to annotate the quality of semantics preservation and syntactic diversity in a 3-point scale question.
We filter the MTurkers by approval rate greater than 97\% and the number of approval greater than 50.
The pay rate is \$0.1 per paraphrase pair.
We do not collect any personal information of MTurkers.

\end{document}